\documentclass{article}

\usepackage[utf8]{inputenc}
\usepackage{todonotes}
\usepackage{url}
\usepackage{natbib}
\usepackage{authblk}

\begin{document}
\renewcommand\Authfont{\scshape}
\renewcommand\Affilfont{\itshape\small}

\title{Ontology Development is Consensus Creation, Not (Merely) Representation}
\date{}
\author[1]{Fabian Neuhaus}
\author[2,3,4]{Janna Hastings}
\affil[1]{Faculty of Computer Science, Otto-von-Guericke University Magdeburg} 
\affil[2]{Faculty of Medicine, University of Zurich}
\affil[3]{School of Medicine, University of St. Gallen}
\affil[4]{Department of Clinical, Educational and Health Psychology,University College London}

\maketitle

\begin{abstract}
Ontology development methodologies emphasise knowledge gathering from domain experts and documentary resources, and knowledge representation using an ontology language such as OWL or FOL. However, working ontologists are often surprised by how challenging and slow it can be to develop ontologies. Here, with a particular emphasis on the sorts of ontologies that are content-heavy and intended to be shared across a community of users (reference ontologies), we propose that a significant and heretofore under-emphasised contributor of challenges during ontology development is the need to create, or bring about, consensus in the face of disagreement. 
For this reason reference ontology development cannot be automated, at least within the limitations of existing AI approaches. Further, for the same reason ontologists are required to have specific social-negotiating skills which are currently lacking in most technical curricula. 
\end{abstract}

\section{Introduction}\label{sec:introduction}

A na\"ive view of ontology development is the following: to develop an ontology, all that one needs to do is to gather the relevant knowledge about a given domain (through studying textual resources or by talking to an expert in that domain) and then formalise it in some appropriate logical language, such as the Web Ontology Language (OWL) or variants of first-order logic (FOL). From this conception, it is often thought to follow that the whole process of ontology development should be amenable to automation. After all, the formalisation of the content of natural language text in a formal language is akin to a translation task from one natural language to another. Since widely available automated natural language translation tools such as DeepL and Google Translate are able to (largely) translate between natural languages, it seems as though the automation of ontology development should be possible. 

In this paper we will argue that this na\"ive view of ontology development is misguided and incomplete -- at least when applied to ontologies that aim to serve a community of users and provide shared definitions for a significant portion of the content of a domain.
It is based on a misunderstanding of what ontologies are, what they contain, and how they are built. In particular, in misconstrues the task of the ontologist as a mere translator, who is tasked with specifying a shared conceptualisation by formalising it in a logical language. 
In actuality, we will argue, the working ontologist is not a polyglot scribe who merely records a pre-established consensus (often referred to as a `shared conceptualisation'). Rather,  her or his main task is to serve as mediator to \textit{bring about} or \textit{create} a shared consensus which seldom pre-exists the ontology development process. This consensus is concerned both with the formal and the informal parts of the ontology content, as will be explained further in the next section.
Usually, it is this mediation, or consensus-creation task that takes up the majority of the time during ontology development, and poses the greatest difficulties, when working to create or extend reference ontologies. 
In contrast, the actual representation of the consensus content in OWL (or some other ontology language) is, typically, a derivative task that requires less effort.  
Although this proposition is unlikely to be news to anyone who is already actively working in ontology development, it is surprisingly not well emphasised in existing ontology development guides and manuals (e.g. \citealp{keet_introduction_2018,arp_building_2015}), and as a result, those who are just starting out in ontology development can be taken by surprise when they discover that this aspect takes so long and is so difficult. Moreover, it is not something that ontologists are typically trained to be able to do, thus they can be unprepared for such challenges the first time they arise. 

But what do we mean by `consensus'? Modern information-rich bodies of knowledge are constantly growing, being added to by the ongoing processes of discovery and knowledge synthesis. 
Yet, they are also subject to disputes and debates, and evidence may be contradictory or fragmented, with different perspectives representing different background theoretical assumptions \citep{balietti_disciplinary_2015,crequit_wasted_2016,hastings_scientific_2021,hastings_mental_2020,elliott_decision_2021}. Ontologies aim to represent the entities that are the subjects of such discourses and debates, and thereby serve an integrative function \citep{phimister_classification_2018,arp_building_2015,hastings_theory_2020,hastings_integrative_2021}. 
However, this does not imply that every aspect of the domain has to be 100\% agreed on before an ontology can be developed for that domain. For most scientific domains, this would set the bar rather too high for their use in practice. 
Rather, it implies that some working agreement needs to be reached about the types of entities to which the discourses are referring, and their core, definitional attributes. 
This working agreement then serves as the foundation for disambiguating when a portion of discourse -- a paper, a study, a report, etc. -- is \textit{about} the same entity, and when it is about different entities, regardless of which specific labels are being used in the texts. 
And it is this core working agreement that we mean by `consensus' in this article. Creating a consensus is a significant aspect of any ontology development project.
As we will discuss in this paper, often this consensus is not intended to be limited to the people actively involved in developing ontology. But, rather, the goal is for the consensus to permeate a wider community, which implicitly adopts the ontology as shared reference point.  

In the remainder of this article, we first give some background information about ontologies and their use, before setting out our case for considering ontology development as a process during which consensus is created or built, rather than the (mere) representation of a pre-existing consensus. Thereafter, we suggest some strategies by which mediation for consensus can be accomplished and some skills the working ontologist needs to have in their toolbox. 
Finally, we discuss some important consequences of this view for the planning of ontology development projects,  for training of ontologists, and for the goal of automating ontology development. In particular, we argue, since automation cannot generate a consensus (at least within current technological capabilities), automatically assembled ontologies face ensuing limitations in applicability.

\section{What ontologies are and how they are used}\label{sec:whatOntologiesAre}

Ontologies are often defined as  \emph{formal, explicit specifications of some shared conceptualisation}.
This definition was proposed by \cite{studer1998} and is a refinement of the original proposal by 
\cite{gruber1993}, who suggested \emph{an explicit specification of a conceptualisation}.\footnote{Both proposed definitions (and their variants) are problematic in the sense that they are not helpful  to provide a better understanding of `ontology'  \citep{neuhausdefinition2018,DBLP:journals/corr/abs-1810-09171}.} 
One interesting difference between the two definitions is Studer's emphasis on \emph{shared} 
  conceptualisation. He argues: `[...] an ontology captures consensual knowledge, that is, it is not private to some individual, but accepted by a group'.

In our opinion, Studer might go a bit far by requiring that every ontology needs to represent knowledge that is accepted by a group consensually.
There are ample use cases for ontologies that are specific to a given task or tailored to a specific software application. In these cases it is conceivable that an ontology reflects only the opinions of a very small group or even one individual. 
However, such examples  would be atypical ontologies in the sense that ontologies very often are designed to address interoperability-enhancing use cases which necessitate adoption among a broader group of stakeholders. Thus, Studer is correct that: 
(a)  Typically, ontologies are intended to be adopted and used by a group of people.  
And (b), typically,  the success of an ontology depends on meeting Studer's consensus-requirement: it needs to represent a consensus by a (significantly sized) group of users.\footnote{
As we will discuss in Section~\ref{sec:creatingconsensus}, often only a small minority of these users are actively involved in the development of the ontology.} 

For the purpose of this paper we are only interested in ontologies that meet Studer's consensus-requirement. 
In particular, this includes 
ontologies that are intended to provide a shared public reference resource for the entirety of a  given domain, e.g., the Gene Ontology \citep{ashburner_gene_2000}, the Chemical Entities of Biological Interest ontology \citep{hastings_chebi_2016}, the Open Energy Ontology \citep{booshehri_introducing_2021}, 
the Semantic Sensor Network Ontology \citep{compton2012ssn} 
and the Financial Industry Business Ontology \citep{bennett_financial_2013}.  Typically, these  ontologies are used for different purposes by their communities of users, including data annotation, integration, aggregation and translation \citep{dessimoz_primer_2017}. It also includes upper level ontologies such as Basic Formal Ontology (BFO) \citep{arp_building_2015} or Descriptive Ontology for Linguistic and Cognitive Engineering (DOLCE) \citep{gangemi2002sweetening}. 
Note that Studer's consensus-requirement does not require an ontology to be publicly shared. E.g., a business ontology that  models the business processes of a  company and is used to integrate information from different departments also captures shared knowledge from a group of people, namely the employees of the company.  

\medskip 
But why is it  so important for  the success of an ontology that it reflects the consensus of a community? 
To understand this, we need to distinguish between two types of knowledge that are represented in ontologies: meaning postulates and falsifiable knowledge. For example, the Pizza Ontology\footnote{The Pizza Ontology is available at \url{https://github.com/owlcs/pizza-ontology}. The Pizza Ontology was developed for training purposes in order to illustrate some aspects of the OWL language and OWL reasoning. We use it in the following as example to illustrate a point. We intend no criticism of the content of the Pizza Ontology.} contains the class `Pizza Margherita' and asserts that such a pizza consists of a pizza base with  mozzarella  and tomato toppings (and no other). 
Together these assertions establish the meaning of `Pizza Margherita' within the context of that particular ontology. 
These meaning postulates reflect a terminological choice of the ontology developers, and, thus, contain no empirical assertion about the world.
If the ontology had also asserted that Pizza Margherita is the most popular kind of pizza in Italy, then this additional information would not alter the meaning of the term `Pizza Margherita' in the ontology. 
Rather it would represent a claim about the world, which could be empirically falsified, e.g., by studying the ingredients and quantity of  pizzas that are sold  in restaurants in Italy. 
Since the representation of knowledge about the world uses terms from the ontology, the truth of the empirically falsifiable assertions depends on the meaning of the terms that are established by the meaning postulates. 
Only after the meaning of term `Pizza Margherita' is established within the context of the ontology, are we able to evaluate claims about Pizza Margherita (in the sense of the ontology) empirically. 
In this sense the representation of empirical knowledge is a secondary task of an ontology, its primary task is to establish a vocabulary for a formal language, which  may be used to express empirical knowledge. 

As written above,  meaning postulates in an ontology reflect a terminological choice. 
As ontology developers are free to decide what the terms in their ontologies are supposed to mean, strictly speaking these meaning postulates cannot be false. 
However, a terminological choice may be at odds with a pre-existing usage in a community. 
E.g., in Italy it is widely agreed upon that a pizza Margherita has three toppings:  mozzarella, tomato and basil leaves. 
Thus, the terminological choice of the Pizza Ontology deviates from an existing consensus of experts. 
This discrepancy would reduce the  chances of the Pizza Ontology being adopted by restaurants in Italy, whose owners might feel strongly about the `correct' use of the term `Margherita'. 
This reduces the usefulness of the ontology for use cases that require its wide adoption (e.g., comparing the data associated with food ordering in restaurants across countries).  
Further, if the Pizza Ontology was adopted by a restaurant in Italy, the discrepancy between the meaning of the term in ontology and the consensual usage would lead to ambiguity and misunderstandings. Inevitably this would cause  complaints by customers who do not receive the pizza they intended to order.   
This example illustrates two major reasons for basing an ontology on a consensus in a community:
It lowers the barrier for adoption in the community. And it makes it easier to enable interoperability (e.g. enabling comparisons between restaurants or between countries), which is one of the major use cases for the development of ontologies.

\smallskip
To summarise the previous paragraphs: \emph{The primary task of an ontology is to  provide the vocabulary for a domain-specific formal language (where the terminological choices for this vocabulary should reflect a consensus of the experts of the domain)}.  
The grammatical rules for that language is provided by some  knowledge representation formalism, e.g., OWL \citep{owl2directSemantics}. 
The semantics of this language  is provided by two sources: 
(i) The logical semantics is provided by the semantics of the logic that the knowledge representation formalism is based on  and the axioms that are included in the ontology. In the case of OWL its (direct) semantics is derived from the model theory of the description logic SROIQ \citep{SROIQ}. It is the semantics that is utilised for automatic reasoning purposes. 
(ii) The intended semantics of the ontology is provided by what the  ontology developers intend the terms to mean (i.e., the `shared conceptualisation'). 
The intended semantics of a term is indicated by choosing a primary label that corresponds to a term in natural language, e.g. `cat'\footnote{Some communities also use IRIs that resemble natural language terms to indicate their intended semantics (e.g., \url{http://www.example.com/Cat}).} and, possibly, by providing other labels as synonyms. The intended semantics is further clarified and documented  by  annotations and comments in the ontology, which contain, for example, natural language definitions, explanations and elucidations. 
Lastly,  in combination with its labels and annotations logical axioms may also contribute to disambiguate and clarify the intended semantics. Hence, axioms play a role in establishing both formal and intended semantics. However, for establishing the intended semantics of a term axioms only play an auxiliary role by constraining possible interpretations, while  labels and annotations moor its semantics by linking it to expressions in some natural language.   

Ontologists with a background in logic tend to emphasise the importance of the logical semantics 
and  tend to focus  on axiomatisation of ontologies. However, even in the case of densely axiomatised ontologies the logical semantics of an ontology is only able to approximate its intended semantics, since isomorphic models are indistinguishable. 
In practice,
most of the ontologies 
that are in actual use currently in domains such as biomedicine are only weakly axiomatised and, thus, whatever is captured by the logical semantics is only a pale shadow of the intended semantics. For example, the intended semantics of the BFO 2.0 OWL ontology is rooted in a philosophical tradition that traces back to Aristotle, expressed in numerous papers and a textbook and documented in rich annotations of the ontology. 
In contrast, the logical semantics of BFO 2.0 OWL is provided by a taxonomy consisting of 34 subsumption axioms and 18 disjointness axioms. These 52 simple axioms obviously do not even capture a minuscule fraction of the intended semantics.\footnote{We are aware that BFO 2.0 OWL contains annotations in CLIF. While these annotations are written in a logical language, they have no impact on the logical semantics on the BFO 2.0 OWL ontology, since they are not interpreted by the OWL direct semantics.} Nevertheless, BFO 2.0 OWL is re-used by a large number of projects.\footnote{\url{https://basic-formal-ontology.org/users.html}}. To consider another example, the ChEBI chemical ontology \citep{hastings_chebi_2016} contains a large, weakly axiomatised hierarchy of molecular entities. For example, ChEBI does not contain logical axioms that  distinguish CHEBI:174059 (1,3-Diphenyl-2-propanone) and CHEBI:34060 (1,3-Diphenylpropane).
 However, since ChEBI contains non-logical content in annotations that represent the molecular structure of both chemicals, the classes are documented in a way that no user of ChEBI would be confused about their intended meaning.
Thus, while the logical semantics of CHEBI:174059 and CHEBI:34060 is indistinguishable, their intended semantics is well specified. 

These examples illustrate that a weak logical axiomatisation does not impede the usefulness of an ontology. (The same point could be made about many widely used ontologies, e.g., the Gene Ontology and SNOMED.) This is the case because in a large number of use cases, ontologies are used as hierarchically ordered controlled vocabularies or dictionaries. 
In these use cases the ontologies provide value  because a community agreed to use a common set of terms with an agreed upon meaning in order to exchange information.  
For this purpose the annotations and comments in the ontology  are much more important than the logical axioms, because they document the intended semantics and, thus, enable a consistent use of the terms in the ontology by different users.

For the ontology development process, these points have two important consequences. Firstly, it follows that a significant portion of the `shared conceptualisation' that is captured by an ontology is not axiomatically specified in the logical formalisation, but is contained in labels, textual annotations and comments associated with the logical axioms. Thus, during the development process the informal part of an ontology requires at least as much attention as the formal part. 
Secondly,  the success of an ontology depends on the fact that the intended semantics of the terms is widely adopted across a community, i.e., reflects a shared consensus in that community. Thus, an ontology development process needs to be organised in a way that supports this goal.

\section{Ontology development process}
The ontology development process has been the subject of a large number of publications, including:   \cite{methonyology1997,gomez2006ontological,keet_introduction_2018,arp_building_2015,uschold1996ontologies,Gruninger1995MethodologyFT,neuhaus2013towards,kendallOntologyEngineering2019,accidentalTaxonomist,
Allemang2021CollaborativeOntologyDevelopment,
abel_knowledge_2005}. 
Each of these publications presents a unique view on ontology development and, thus, any generalisation is difficult. However, summarising the literature in very broad strokes,  ontology development can be characterised as a process that  includes the following core steps:
\begin{itemize}
    \item Identify and scope the relevant terminology for the domain: Scanning the literature, identifying pre-existing standards, ontologies and controlled vocabularies, as well as data and schemas associated with domain-relevant databases, and consulting with domain experts
    \item Consult dictionaries and domain experts in order to associate the candidate terminology with candidate definitions
    \item Placing each candidate entity hierarchically within the overarching taxonomic structure of the ontology
    \item Adding relevant logical axioms where possible, and 
    \item Aligning the resulting content to an upper level ontology
\end{itemize}

Some of these guides and methodologies do refer to the involvement of a wide community of users in the development process. For example, involvement of users in ontology development is an OBO Foundry \citep{smith_obo_2007} principle, and formal consultation of expert stakeholders is recommended in \citep{wright_ontologies_2020,norris_why_2021}. 
However, none of the ontology development guides explicitly highlight the need for the \textit{creation} of a shared consensus about the representation of entities in a domain \textit{where none yet exists}. 
As a result, people who are new to ontology engineering and, thus, are consulting these guides 
are led to believe that  such a consensus is usually pre-existing, and the work involved in ontology development amounts to the determination and representation of that consensus.  

Indeed, the assumption of consensus in the domain is seen by some as an explicit precondition for the development of ontologies. For example, \citep{arp_building_2015} write: 
\begin{quote}
"Very occasionally, ontologies may need to be developed to support research in areas still subject to dispute among different groups of scientists and thus not belonging to established science. (Recall, again, the case of the “Higgs boson.”) We prefer to see such ontologies as provisional in nature, to be promoted to the ranks of ontologies proper only when the disputes in question have been settled. The methods for the creation of such provisional ontologies will then be essentially the same as those outlined here, but will apply the process of term selection not to established textbooks but, for example, to journal articles produced by some subset of the disputing partners. The results of such provisional ontology development will then also be provisional. They will be able to be added to existing ontology content and treated like other ontologies only when the disputes in question have been resolved." (page 81). 
\end{quote}

However, while in established sciences in the sense of \citet{arp_building_2015} there are indeed no major controversies and therefore there is a general agreement on the big picture, in our experience there are still a lot of competing views when one looks at the details. Hence, while there is a generally shared view among experts, even in `settled science' the agreement  is typically not at the level of detail that is needed for ontology development. 

This is because in most if not all domains, there are many experts with their own points of view, theoretical assumptions, and scientific traditions. These lead to different conceptualisations of the domain, different terminologies and different semantics of shared terms. Often the domain experts are not even aware of these differences, because humans are very accustomed to negotiating these differences when they communicate with each other. However, they become apparent during the ontology development process, because they are an obstacle for agreement on shared definitions and axioms. 
These differences are exemplified by the wide range of different definitions in use for any given entity. 

Let's consider the term `cell' as an example. 
The theory that organisms are made up from cells is established science in the sense of \citep{arp_building_2015}. 
Thus, the broad-stroke definition of `cell' and their existence is not disputed in molecular and cellular biology. Further, cells are material objects that may be studied directly by observation. Terms for these kind of entities tend to be less controversial than, for example, terms for mental phenomena or social constructions (e.g., `imagination' or `gender'.) Hence, one might expect that biologists would agree on the definition of `cell'.

Nevertheless, different definitions of `cell' are in use. Here are some examples: 
\begin{itemize}
    \item \textit{Dictionary definition \citep{lexico_cell}:} `The smallest structural and functional unit of an organism, which is typically microscopic and consists of cytoplasm and a nucleus enclosed in a membrane.' 
    \item \textit{Gene Ontology definition \citep{ashburner_gene_2000} (now obsolete as superseded by the cell ontology definition given below):} `The basic structural and functional unit of all organisms. Includes the plasma membrane and any external encapsulating structures such as the cell wall and cell envelope.' 
    \item \textit{Common Anatomy Reference Ontology definition \citep{haendel2008caro}:} `An anatomical structure that has as its parts a maximally connected cell compartment surrounded by a plasma membrane.'
    \item \textit{Cell Ontology definition \citep{meehan_logical_2011,diehl_hematopoietic_2011,diehl_cell_2016}:} `A material entity of anatomical origin (part of or deriving from an organism) that has as its parts a maximally connected cell compartment surrounded by a plasma membrane.'
\end{itemize}

The differences between these definitions are not huge, and in broad outline they are conveying the same information. But the precise details differ: for example, the Cell Ontology definition encompasses cells both \textit{in vivo} and \textit{in vitro} hence refers to the possibility that the cell is part of an organism \textit{or} derived from one. 
Note that the latter three definitions are already the results of attempts of ontology developers to forge a consensus. 
Since one of the authors was personally involved in the development of CARO, we can report that its definition of `cell' was the result of several hours of intense discussion between a group of experts -- and that the resulting definition was considered an unsatisfactory compromise by some.  
This example illustrates that even under the best of circumstances (namely, a widely used term from a well-established, non-controversial theory in empirical science), an ontologist cannot rely on the existence of a pre-existing consensus in the community, which just needs to be recorded in the ontology.

While, as we mentioned above, this aspect has largely been under-emphasised in existing guides to ontology development, there are two prior publications, in which it is explicitly recognised that some form of negotiation between experts will be required \citep{aschoff_knowledge_2004, aschoff_knowledge_2004b}. Aschoff et al. explicitly suggest that ontology development should take place as a group activity amongst experts, \textit{as mediated by an ontologist}. They see the development process as involving three separate steps: 
\begin{itemize}
    \item \textit{Generation:} Brainstorming of terms by domain experts (no discussion / critique)
    \item \textit{Explication:} Individual domain experts create ad-hoc ontologies; prioritisation of entities
    \item \textit{Integration:} Proposals are integrated by domain experts; the  ontology expert serves as a neutral mediator
\end{itemize}

In this workflow, the  definitions of terms and any other content of the ontology are solely the result of a  negotiation process between  domain experts. 
The ontology expert facilitates this process, and ensures that in these negotiations the interests of all of the participants are considered equally. 
Thus, according to this view,  the role of the ontology expert is 
restricted to a neutral mediator; she enables communication between experts, but does not influence the content of the resulting ontology.

By pointing out the fact that consensus creation is an essential part of ontology development, Aschoff et al. made an important contribution,  which in our opinion has unfortunately not received the attention it deserves from within the ontology development community. 
However, from our point of view  ontology experts have a more active role to play than being neutral honest  brokers, who mediate the discussion between domain experts. 
Firstly, because the ontology experts have valuable expertise that the domain experts typically lack: 
They can use their ontological expertise to ensure that the architecture of the ontology is sound and help the domain experts to formulate ontologically and logically sound definitions. 
Further, they have the logical nous to support the domain experts in the development of axioms in a formal language like OWL, and are also aware of the limits of these formal languages. In addition, they can offer guidance on how to set up the ontology development process (e.g.,  workflow, software tools, deployment process, copyright license). 
Secondly, we believe that the consensus that the process of ontology development brings about should not merely be a 
 consensus among `the people in the room', but rather, should become a consensus among all the potential users of the ontology. Thus, for a reference ontology the consensus should encompass the community involved in the domain as a whole. Hence, the ontologist does not just need to mediate between those views represented by the experts within the development team, but, typically, also needs to ensure that the ontology development process is designed to engage with a wider community.

In summary, the work of getting the actual knowledge that needs to be represented in an ontology is always more complex than just acquiring an established consensus from domain experts. Rather, it becomes necessary for the ontologist to mediate the discussion process which brings about a consensus. However, the role of an ontologist is not that of a neutral broker, but he is actively engaged in shaping the outcome by providing expertise that the domain experts lack. We contend that this is a big step that is missing in the existing ontology development guides: the process of \emph{guided mediation} to arrive at a consensus for the definitions, axioms, and annotations of the entities included in the ontology.

\section{Creating consensus}\label{sec:creatingconsensus}

Consensus creation can be thought of as an additional, under-appreciated step in the ontology development process, which takes place after experts, dictionaries and other terminologies have been consulted. However, consensus creation can also be thought of not as an additional and discrete step but rather as the output of the entire ontology development process. From this perspective the objective of the ontology development is to bring about a  consensus view on 
a particular domain, which is documented by the ontology.

As discussed in Section~\ref{sec:introduction}, by `consensus' we do not imply that all issues are settled and that arguments have ceased. Rather, we are talking about a working agreement about the types of entities to which the discourses in a given domain are referring, and their most important relationships. This consensus is reflected by the meaning postulates in the ontology that determine both the logical and the intended semantics of its vocabulary. 
Consensus creation involves two distinct steps: reaching agreement between domain experts that are actively engaged in ontology development, and convincing the intended users who were not actively involved during the development process to adopt the ontology.  
Thus, we distinguish two levels of consensus: a micro-level consensus between the ontology developers, and a macro-level consensus by  the wider community that the ontology is intended to serve.

\subsection{Creating micro-level consensus}
The process of building the consensus that is represented by an ontology necessarily includes everyone involved having the chance to have their voices heard. This part of the process creates the attitudes that are needed for the success of the negotiation. After each person has had their position heard and taken seriously, as well as understood where the points of disagreement actually are, it is much more likely that they will then accept the outcome of these negotiation, mediation and elaboration processes, which we term consensus building.

Thus, one major task of ontologists is to establish an inclusive participatory development process and to serve as mediator when the domain experts disagree on a given subject.
In addition, we see the role of ontology experts as active guides for domain experts during the ontology development process.
They  help to formulate requirements and set the scope, shape the architecture of the ontology and its broad ontological (and upper-level) commitments.
Further, they establish a good development workflow that involves enough time and opportunity for consensus creation, and serve to guide the ongoing development process towards an ontologically sound consensus. 
This way, the ontology will truly come to embody and represent a shared consensus. However, this will not be a consensus that pre-existed the ontology development, but, rather, a consensus that is forged during the inclusive discussions and debates that are part of ontology development.

Developing an ontology involves achieving agreement on three aspects: (a) the scope of the ontology, (b) the entities that exist and their relationships (i.e., the ontological analysis), and (c) the terminology that is used to denote them. 
During later stages of the development of an ontology, the ontology itself provides a shared terminological and conceptual foundation that enables discussions between ontology developers about these aspects. However, 
at the start of an ontology development project, typically there is no consensus on any of these aspects. Thus, when   domain experts disagree, it is often difficult to pinpoint whether the disagreement is about scope, ontological analysis or about terminology -- or whether all aspects are intertwined.

Hence, an important first step in an ontology development project is to create a shared frame of reference that may be used to bootstrap the ontology development process.  
One strategy that we have applied successfully in the past might be called `explain the project to the na\"ive ontologist'. For this technique the domain experts are required to explain to the ontologist collectively how they are planning to use the ontology and what the entities are. 
During the conversation with the domain experts the ontologist draws -- visibly for everyone involved -- a diagram representing the  kinds of entities and their relations that the ontology needs to cover to support their goals. 
One advantage of this approach is that most people like to talk about their work, so it is easy to engage the domain experts. 
Further, since the experts are addressing somebody who lacks any of the background knowledge that they usually take for granted, they are forced to use more basic language and to explain their perspective in much more detail than they would if talking to their peers.  
This enables the other domain experts in the room to get a better understanding of their respective perspectives. Since the diagram provides a visual model of the domain, it often helps to elicit a reaction from the other experts in the room. 
The fact that the ontologist draws the diagram has two benefits: Firstly, she can ensure that the diagram is ontologically sound (e.g., resolve ambiguities between processes and functions). Secondly, since the ontologist draws the diagram, he `owns' all of its errors. Hence, there is a lower barrier for domain experts to correct the model, than there would be if the diagram were drawn by another expert.
In our experience, this exercise often leads to extensive discussions between domain experts. Even if the ontologist might not be able to follow the details of this discussion, these discussions are important, because they help to establish the shared framework, which provides the foundation for further ontological work. 
These discussions might lead to the recognition of ambiguities or differences of opinions. Unless resolving these is required to determine the scope of the ontology, it is often best to shelve contentious issues at this stage. 
The output of this work is an informal, very general and incomplete model of the domain, which provides the starting point for the ontology. Since it covers the kind of entities that are relevant for the development of the ontology, it fixes its scope (at least for some time). And it provides a shared frame of reference and a shared terminology for future discussions.

After an initial version of the ontology is established, we recommend establishing a workflow for the continued development. 
Since one goal of the development process is to support the building of consensus, the workflow should  be designed to be participatory and transparent, and decisions should be documented. This might be realised by a workflow involving the following points: 
\begin{itemize}
    \item Uncontested extensions or changes may be agreed on  offline; e.g. via discussions on GitHub issues or some other transparent medium. 
    \item Contested extensions or changes are be triaged or prioritised. Essential, prioritised entities are taken to a mediated group discussion. 
    \item In these discussions the ontologist  guides the experts towards an ontologically sound consensus, which may involve mediating between different perspectives. 
	\item There need to be some rules about decision-making if consensus cannot be achieved. However, in general, it is desirable to aim for unanimity. And if no consensus may be reached on a proposed extension (typically, the definition of an entity), it is usually better to not include it. 
\end{itemize}

The mediation role of the ontologist during the group discussions  involves several different objectives. The ontologist needs to ensure that all participants are able (and feel welcome) to present their perspectives. Further, the ontologist needs to lead and structure the discussion in a way that contributes to the resolution of differences of opinion between experts.  \cite{aschoff_knowledge_2004b} present several mediation techniques for that purpose. 
However, as mentioned above, the role of the ontologist is not just to act as a neutral mediator, but to ensure ontological soundness. 
 Three techniques that are particularly useful for this purpose are \emph{asking na\"ive questions},   \emph{summarising} and \emph{paraphrasing}.
Asking na\"ive questions is basically the same technique that we described above, but instead of asking the domain experts to explain the whole project, the questions are focused on the given contentious issue. By explaining their proposal to a third, na\"ive party, the experts are forced to be more explicit, which helps them to better understand each other.
By summarising the ontologist tries to distill the results of a conversation in some ontologically sound definition (or some axioms or annotations). This enables her to establish progress that has been made, and refocus the conversation on open questions. 
 By paraphrasing a proposed definition  the ontologist tries to capture the essence of the proposal, but formulates in a way that fits into the context of the ontology, ensures logical quality (e.g., avoidance of circular definitions), and is  coherent with the design of the ontology (e.g., analysis of the domain consistent with a given upper ontology).  
In the case of competing proposals,  paraphrasing the different options in an ontologically sound way will typically clarify the nature of the disagreement; which is often a first step towards resolving the differences of opinion.  If all members of the ontology development team are sufficiently proficient in a formal language (like OWL), paraphrasing might include both natural language definitions and logical axioms. If this is not the case, it is better to also present  the content of the future logical axioms in natural language, ideally in a way that is sufficiently explicit and unambiguous that a translation into a formal language is straight forward. 

One major obstacle towards consensus building is the confusion between ontological and terminological questions. While it is essential to reach consensus on the entities and their relationships that are represented by the ontology, terminological differences can be accommodated more easily in the ontology. For that reason alone it is always beneficial to reach agreement on the ontological questions first and tackle terminological concerns later. Unfortunately, most people are not trained to distinguish between ontological and terminological differences. Thus, the ontologist often needs to intervene to separate discussions about \textit{terms} (labels) from discussions about \textit{entities} (e.g., classes, relations).

 Methods for resolving disagreements during ontology development that relate solely to the use of terms have featured in previous approaches to knowledge acquisition and ontology development. For example, \cite{uschold_towards_1995} suggest the following  
approach for dealing with ambiguity in particular. But it is applicable more generally to enforce a separation of  ontological and terminological discussions: 
 \begin{itemize}
     \item If a term has many possible meanings, it may be helpful to suspend use of that term; it is too ambiguous.
     \item The underlying ideas may then be clarified by carefully defining  each concept  using  as  few  technical  terms  as  possible, or only those whose meaning is agreed. 
     \item It can be helpful to give these definitions meaningless labels such as x1, x2, x3 etc. so they can be conveniently referred to during discussions. 
     \item Determine  which,  if  any  of  these  concepts  are  important  enough  to  be  in  the  ontology, and finally,
     \item Choose  an appropriately specific  term  for  that  concept  (avoiding  the  original  ambiguous  term,  if  at  all possible). 
 \end{itemize}
 
Meaningless  labels are not just useful as a mediation device, but we recommend the use of  meaningless alphanumeric identifiers 
for the entities in the ontology (in line with best practice recommendations by the OBO Foundry and others, e.g. \citep{mcmurry_identifiers_2017}). One advantage of this approach is that it makes the separation between the ontological and terminological aspects of an ontology explicit. In particular, it helps to ease discussions about the `right term' for a given entity, because diverging terminologies may be represented by alternative labels of the same entity (without giving priority to any of them).  
Polysemy may be represented by using the same term as secondary label for different entities in the domain. However, in order to be able to denote an entity unambiguously, each entity should be associated with at least one unambiguous human readable term. 

Since diverging terminological choices can be accommodated within the ontology, developing an ontology does not require the domain experts to always agree to use the same terminology for their domain. However, it requires consensus on the denotation of any given term in the context of the ontology.  
 
Once consensus has been reached about core entities and relationships for some aspect of the ontology, in some cases that consensus can also be extended to a range of additional entities that follow the same \textit{pattern}, via the implementation of ontology design patterns \citep{gangemi_ontology_2009,shimizu2021modular} or templates \citep{skj230_chapter_2021,jackson_robot_2019}. For example, the representation of metabolic processes in the Gene Ontology is structured by a template in which metabolic processes are defined by the small molecular metabolites that participate in them \citep{hill_dovetailing_2013}. The relevant template specifies that for a given metabolite M, the class of metabolic processes involved has the name \textit{M metabolic process}, synonym \textit{M metabolism},  and a relationship `has participant' to M.

\subsection{Creating macro-level consensus}
It is one task to lead the `people in the room' to an agreement; it is a different task to convince intended users, who are not directly involved in the development of the ontology, that they should adopt it. The amount of effort that is required to achieve macro-level consensus depends on the given ontology development project. 
If, for example, the ontology has been developed for a company, representatives of all affected departments are participating in the development process, and the company management enforces the utilisation of the ontology, then macro-level consensus is not an issue. In contrast, in the case of a reference ontology that is intended to be a shared resource for a large domain (e.g., in the case of the Gene Ontology: cellular components, molecular functions, and biological processes related to gene products in all species),  the ontology development team will necessarily only involve a small fraction  of its intended users. In this case the ontology development process needs to be designed in a way that encourages potential users outside the development team to adopt the ontology. 

One requirement for the wide adoption of an ontology is extensibility, because often new users need to cover some aspect of the domain in more detail than it is already represented in the ontology. 
Extensibility requires that the ontology developers should not include incorrect annotations or wrong axioms, even if it seems like a justified  simplification at the time.\footnote{In our opinion, this is one of the most important lessons from the ontology development methodology called `ontological realism'. This is a somewhat misleading name, since the methodology is independent from an endorsement of the metaphysical doctrine of ontological realism concerning the existence of universals \citep{smith2010ontological}.} 
Because every time one includes some axiom in an ontology that is only true in a specific context, one limits the usage of the ontology to that context and reduces its extensibility and, thus, its potential of adoption. 
E.g., including \emph{All Swans are white} in an ontology might seem a safe simplification for the ontology developers who are only concerned with birds in Europe, but it closes the opportunity for its reuse in Australia. 
For this reason it is preferable to formulate the definitions and axioms in an ontology in a way that is not tailored to the particular use case that prompted the development of the ontology. Often this goal may be simply achieved by introducing a more specific term (e.g., \emph{All Mute Swans are white}). If this solution is not possible, instead of including wrong assertions in an ontology, it is better to simplify by providing  incomplete information or by substituting specific terms by more generic terms.

Beyond the content of the ontology, it is important to organise the ontology development process in a way that the larger community perceives the ontology as a shared resource. 
 In practice, the ontologist should encourage: 
\begin{itemize}
    \item that the development process is open and participatory, and the outputs are accompanied by suitably permissive licenses,
    \item that the decisions taken in the development of the ontology are transparent,
    \item that there is extensive documentation which is also accessible to non-experts and those who have not been involved in the development to date,
    \item that there is a low barrier for community members to report problems or request changes to the ontology, to which the ontology developers and maintainers are responsive, 
	\item that the development team is open for new members who would like to join,	and 
    \item that there are efforts to actively engage the wider community (e.g., through the appointment of a steering committee and the organisation of tutorials at relevant conferences). 
\end{itemize}  

Indeed, many of these macro-level participatory consensus-building aspects are reflected in the principles of the OBO Foundry \citep{smith_obo_2007}, which has as its objective the steering of a community of practice for life sciences reference ontology development. 

One major benefit of adopting  an ontology is that it enables the integration of ones data with existing datasets that already use the ontology. Thus, an additional way to encourage  the adoption of an ontology is the publication of resources that utilise its terms. Since each additional resource increases the potential for data integration for all users of the ontology, this may lead to a positive feedback loop that establishes the ontology as the \textit{de facto} terminological standard within a given domain.     

\medskip
In summary, the role of the ontologist is more than to be a capturer of a pre-existing consensus, and more than to be a disinterested mediator or broker of consensus between domain experts. The role of the ontologist is to \textit{bring about} the consensus by mediating between domain experts and by applying the requirements of logical and ontological coherence, which pose additional constraints on the evolution of the definitions that interact with the requirement for negotiating between different perspectives that different domain experts may have. Taken together, a consensus that grows out of considering a variety of domain experts' perspectives is likely to be significantly more robust and well-thought-out than one developed without. One important aspect of the quality of an ontology is its extensibility, because it enables the adoption of the ontology by users who are not directly involved in the development of the ontology. 
If that happens, the ontology takes the step from representing a consensus position shared among its developers to becoming an integrating terminological resource for a larger community.

\section{Implications}
As we have shown, ontology development does not consist of just representing a `shared conceptualisation', but is -- to a large degree -- the process of creating it (in a ontologically informed way). If done properly, ontology development leads to 
a newly created, logically consistent and ontologically sound consensus, which integrates divergent views and is adopted by  different stakeholders of a given community. Thus, the ontology development leads to something, which did not exist before the ontology development process started. And, often, this consensus itself is of significant benefit to the community   -- independently of any applications of the ontology in information systems.  

In this section we will discuss some consequences of this perspective on ontology development.

\subsection{Implications for the automatic generation of ontologies}
Ontology development is both expensive and time-intensive. 
Ontology learning is a field that tries to address these issues by developing methods to automate the generation of ontologies.   
\cite{asim2018survey}  analysed over 140  papers on ontology learning in a recent survey. They summarise the task of ontology learning as follows:
\begin{small}
\begin{quote} 
\emph{[...] instead of handcrafting ontologies, research trend is now shifting toward automatic ontology learning. Whenever an author writes something in the form of text, he is actually doing it by following a domain model in his mind. He knows the meanings behind various concepts of particular domain, and then using that model, he transfers some of that domain information in text, both implicitly and explicitly.
Ontology learning is a reverse process as domain model is reconstructed from input text [...] }
\end{quote}
\end{small}

The quote is somewhat misleading, since ontology learning techniques are not typically applied to a single document, but to corpora of unstructured domain-specific documents. {These corpora vary in size,  \cite{asim2018survey} mention a list of popular datasets ranging between 307 and 348,566 documents.} The corpora are analysed by a  broad range of approaches combining linguistic,  statistical and logical techniques in order to extract concepts, relations, and axioms. 

\smallskip  
While the output of an automated ontology learning tool might formally look like an ontology (e.g., it might be a syntactically correct OWL file), 
this is not sufficient to qualify as an ontology. 
Because, as we discussed in Section~\ref{sec:whatOntologiesAre}, an ontology  provides
a vocabulary for a domain-specific formal language, where the terminological choices for this vocabulary are supposed to reflect a consensus of the experts of the domain. 
This is hard, if not impossible, to achieve by automatic ontology learning processes alone for two major reasons.   

Firstly, as the quote above illustrates, an underlying assumption of ontology learning approaches is that a `domain model' may be reconstructed from a given text. But in order to reflect not the position of a single person but to represent a community, it is necessary to look at many documents from a variety of authors.  
However, as we have seen above, typically experts in a given domain  disagree  to a significant degree on what entities exist, how they relate to each other, and on the terms that are used to denote entities and relations. This lack of consensus will be reflected in the literature. 
Thus, there is no single `domain model' or `shared conceptualisation' that may be extracted from a text corpus of any significant size. 
 Rather, any automatic ontology learning tool will encounter the bewildering ambiguities,  inconsistencies and differences of opinion 
  that human ontology developers face when they develop an ontology. These divergent perspectives cannot be resolved by the linguistic, statistical or logical methods that are prevalent in ontology learning. Ontology developers use approaches like the ones we discussed in Section~\ref{sec:creatingconsensus}. It seems unlikely that these may be replaced by automated processes anytime soon.  

Secondly, even a perfect ontology learning tool may only replicate the \emph{result} of ontology development, but never the \emph{process}. And that process is important for the adoption of the ontology. Let's consider an analogy: Assume  we have created some AI that is able to predict election results perfectly. Does that mean we could omit the vote and just use the prediction by the AI as election result instead? Obviously not. 
The reason is that the process of voting is important: it is important for people to have their chance to have their voice heard, even and especially if they are on the losing side. If we were to replace the vote by some predictions by some (supposedly) infallible AI, the `election' would have no legitimacy, because people would  not accept the result. 
The same holds for ontology development. By being involved in the ontology development process, domain experts develop a feeling of ownership. Further, because they are able to  participate and shape the result, they are willing to adopt a consensus position, 
even if that consensus diverges from their original point of view.  If one excludes the stakeholders of intended user community, the relevant people will feel disenfranchised and not support the adoption of the ontology. 
And since an ontology is only successful if it is adopted by its 
 intended community, a participatory development process is essential for the success of the ontology.
Thus, even in the theoretical case that an ontology learning tool could generate an  ontology of the same quality as a manual development process, the automatic generation of content is by definition not a participatory process. Thus, 
it would be difficult to gather support for the adoption of that ontology.   
\medskip

We want to stress that we understand the economic argument for (semi-)automation of ontology development. 
It is one of our main motivations for working on techniques for automatically extending existing ontologies\citep{interpretable_ontology_extension_2022, memarianiOntoExtension2021, hastings2021LearningChem}.
Further, we see a lot of value in  ontology learning techniques for the development of semi-automatic tools that {support} ontology developers in their tasks. 
However, the vision of ontology learning as formulated by \cite{asim2018survey}, according to which handcrafted ontologies would be replaced by completely automatically created ontologies, is based on a fundamental misunderstanding of what ontologies are and what they represent.

\subsection{Implications for planning an ontology development project}

Since building a consensus is a significant part of ontology development, 
this aspect should be considered during planning of an ontology project. 
In particular, the amount of effort and time it will take to develop an ontology will depend on the degree of pre-existing consensus in the relevant community with respect to the domain of the ontology. 
For example, since there is more theoretical diversity in the social sciences than there is in chemistry, the development of an ontology in social sciences will, typically, require more effort than the development of an ontology of a comparable size in chemistry.

Further, there are consequences for the selection of the members of the development team. The team should include domain experts that are representative of different intended user groups and their views on the domain.\footnote{A similar point is made by \cite{shimizu2021modular}.}  
(E.g., in the case of a business ontology that would be the different departments that are supposed to use the ontology. In the case of a reference ontology it would be different research communities that are intending to share data with the help of the ontology.) 
Ideally, the domain experts involved are both flexible enough to accept the compromises that are usually required in order to negotiate the consensus, and are able to explain the ontology (and possibly contentious design decisions) to other members of their community. 
In addition, it is desirable that the team includes  ontological expertise (e.g., is aware of useful ontological distinctions and recurring design decisions), some expertise in a relevant representation language (e.g., OWL) and its underlying logic, and the expertise of mediating differences of opinion and leading a discussion towards a shared consensus.  Often the latter are provided by a dedicated ontology expert.

In the case of reference ontologies that are intended to be used as public resources, the members of the development team will only represent a small number of potential users. Hence, developers of a reference ontology need to address the question, how to convince people, who have not personally been engaged in the  development of an ontology, to adopt the consensus it represents. This is mainly a question of achieving critical mass, because the more datasets and tools use a particular ontology, 
the more incentives there are for third parties to also adopt the ontology. But any project that plans to develop a reference ontology needs to consider how to achieve that critical mass. 
In Section~\ref{sec:creatingconsensus} we discussed several   measures that increase the potential of an ontology to be adopted by a wider community; e.g., by publishing the ontology under a permissive copyright license and by organising an open, participatory, and transparent  development process  that engages the stakeholders of the larger community.

\subsection{Implications for the training (and hiring) of ontologists}
During the  Ontology Summit 2010, leading  ontology experts discussed the training and education of future ontologists. 
In the resulting document the authors identify core knowledge and core skills that are required for an ontologist to be able to develop ontologies successfully \citep{neuhaus2011creating}.  Of the 15 core skills that were identified, 14 are technical (e.g., the ability to conduct an ontological analysis or choose an appropriate representation language). Only one of the core skills is social, namely the ability to work in teams.

Since creating a consensus is major aspect of ontology development, in our opinion there are in fact several additional social skills  that are essential for any ontologist (or anybody who leads an ontology development project).   These include: 
\begin{itemize}
    \item  The ability to create an environment for fruitful discussions between members of the development team. This needs to take into account that members may have different social status or different ways to communicate their points. 
    \item The ability to support  domain experts in formulating their (often implicit) knowledge in the form of ontologically sound definitions and axioms. Often this  requires asking the right questions, grasping the essence of often imperfectly worded ideas, and paraphrasing them.
    \item The ability to quickly understand conflicting views, to identify the points of contention, and to present them to the domain experts. 
    \item After the points of disagreements have been made transparent, the ability to lead a discussion and guide the participants towards a consensus. 
\end{itemize}

Note that we do not question the importance of the technical skills that are mentioned by \cite{neuhaus2011creating}, but we do argue that the authors got the balance wrong. 
In our experience, given the choice between somebody with brilliant technical skills and few social skills or somebody with moderate technical  and good social skills, we would  hire the second  person to lead an ontology project. 

\section{Conclusions}

A pre-existing `shared conceptualisation' is a rare luxury. Thus, the job of an ontologist is not merely to represent content `as given to them'. 
Rather, a significant part of the work of an ontologist is to help to create what will then be represented, namely, an ontologically coherent  view on the domain that is agreed upon by the domain experts. This consensus may itself be a major contribution to the field. 

In this paper we discussed approaches for achieving such a consensus between the ontology developers and for communicating it to a wider community. These approaches include both a `macro-level' dimension of transparency and involvement of the wider community, and a `micro-level' dimension involving strategies for identifying points of disagreement and bringing about consensus through elaboration, clarification and negotiation.    

This aspect of ontology development has so far been, in our opinion, not fully appreciated in the ontology development literature, and it has several important implications: (a) it severely limits the potential for fully automating the process of ontology development; (b) the level of (dis)agreement within a given community impacts the amount of resources required for an ontology development project and (c) the skill set that is required from an ontologist who leads an ontology development project is not only technical, but substantially social.

\nocite{*}
\bibliographystyle{ios2-nameyear}  
\bibliography{preprint}        

\end{document}